\documentclass[conference]{IEEEtran}
\IEEEoverridecommandlockouts
\usepackage{cite}
\usepackage{amsmath,amssymb,amsfonts}
\usepackage{graphicx}
\usepackage{textcomp}
\usepackage{xcolor}
\usepackage{multirow}
\usepackage{algorithm}
\usepackage{algorithmicx}
\usepackage{algpseudocode}
\usepackage{array}
\usepackage{textcomp}
\usepackage{stfloats}
\usepackage{url}
\usepackage{verbatim}
\usepackage{graphicx}
\usepackage{cite}
\usepackage{amsmath}
\usepackage{amssymb}
\usepackage{booktabs}
\usepackage{times}
\usepackage{epsfig}

\usepackage{multirow}
\usepackage{makecell}
\usepackage[table]{xcolor}
\usepackage{subcaption}
\usepackage{enumerate}
\usepackage{inconsolata}
\usepackage{diagbox}
\usepackage{tcolorbox}
\tcbuselibrary{breakable}
\usepackage{subfloat}
\usepackage{caption}
\usepackage{tabularx}
\definecolor{mygray}{gray}{0.87}
\usepackage[pagebackref,breaklinks,colorlinks]{hyperref}
\usepackage[capitalize]{cleveref}

\def\BibTeX{{\rm B\kern-.05em{\sc i\kern-.025em b}\kern-.08em
    T\kern-.1667em\lower.7ex\hbox{E}\kern-.125emX}}
\begin{document}

\title{Progressive Video Condensation with MLLM Agent\\ for Long-form Video Understanding}

\author{
Yufei Yin\quad
Yuchen Xing\quad
Qianke Meng\quad
Minghao Chen\quad
Yan Yang\quad
Zhou Yu\thanks{Zhou Yu is the corresponding author} \quad\\
\normalsize Zhejiang Key Laboratory of Space Information Sensing and Transmission,\\
\normalsize Hangzhou Dianzi University, China \\
{\fontfamily{pcr}\selectfont \small \{yinyf, mqk, chenminghao, djj, shaozw, yuz\}@hdu.edu.cn}\\
}

\maketitle

\begin{abstract}

Understanding long videos requires extracting query-relevant information from long sequences under tight compute budgets. Existing text-then-LLM pipelines lose fine-grained visual cues, while video-based multimodal large language models (MLLMs) can keep visual details but are too frame-hungry and computationally expensive.
In this work, we aim to harness MLLMs for efficient video understanding. We propose ProVCA, a progressive video condensation agent that iteratively locates key video frames at multiple granularities. ProVCA first adopts a segment localization module to identify the video segment relevant to the query, then a snippet selection module to select important snippets based on similarity, and finally a keyframe refinement module to pinpoint specific keyframes in those snippets. By progressively narrowing the scope from coarse segments to fine frames, ProVCA identifies a small set of keyframes for MLLM-based reasoning.
ProVCA achieves state-of-the-art zero-shot accuracies of 69.3\% on EgoSchema, 80.5\% on NExT-QA, and 77.7\% on IntentQA, while using fewer frames than previous training-free methods.
\end{abstract}

\begin{IEEEkeywords}
Long-form video understanding, multi-modal large language model agent.
\end{IEEEkeywords}

\section{Introduction}
\label{sec:intro}

Video understanding, which interprets spatiotemporal information to analyze events, actions, and relationships, is challenging due to the high dimensionality and temporal length of videos. Multimodal Large language models (MLLMs) have demonstrated strong long-context reasoning capabilities \cite{achiam2023gpt}, motivating their use in video understanding tasks \cite{zhang2023simple, wang2024videotree}. Recent advances in multimodal modeling and instruction-based learning enable LLMs and MLLMs to handle text, images, and videos within extended contexts, making them promising tools for video reasoning.

\begin{figure}
    \begin{center}
        \includegraphics[width=1\linewidth]{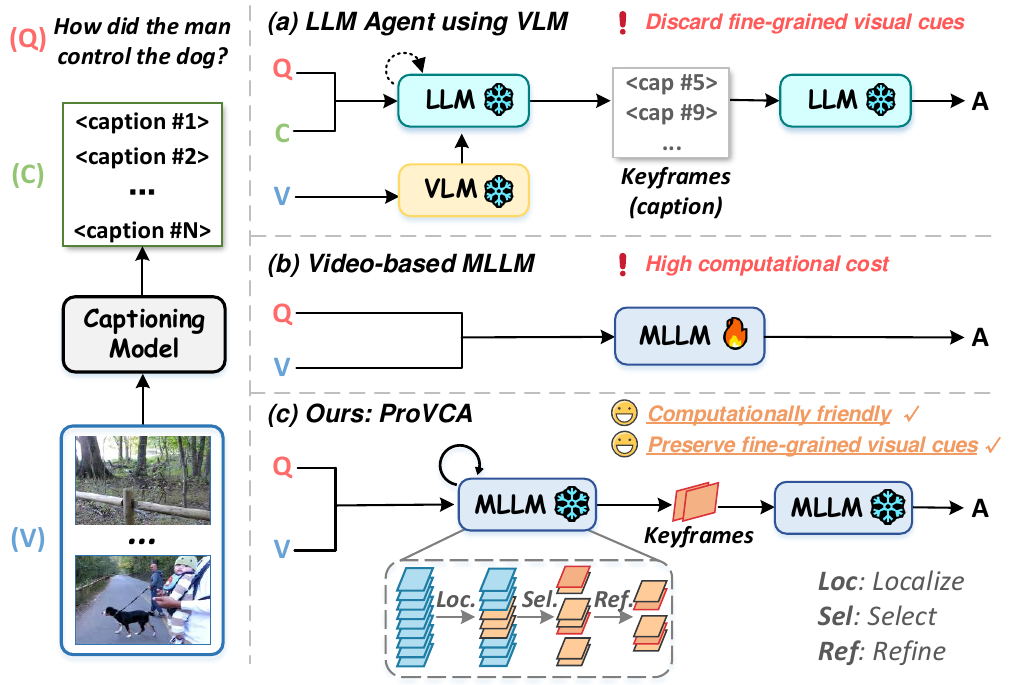}
         \caption{Conceptual comparisons of three Video understanding methods based on LLM (MLLM). Existing approaches either (a) directly reason with pre-trained video-based MLLMs at high computational cost, or (b) use LLM agents that convert frames to captions and rely on VLMs (e.g., CLIP) to filter key frames, discarding fine-grained visual cues. In contrast, our ProVCA (c) progressively condenses videos by using an MLLM to narrow down query-relevant content from coarse segments to fine frames, then feeds the selected key frames into the MLLM to generate the final answer.}
        \label{fig:intro}
    \end{center}
    \vspace{-0.57cm}
\end{figure}

One mainstream video reasoning approach \cite{wang2024videotree,wang2025videoagent} (\cref{fig:intro}~(a)) constructs LLM-based agents following a text-then-LLM pipeline, where video frames are first converted into textual descriptions using a visual captioning model, and the generated captions are then passed to LLMs (e.g., GPT-4 \cite{achiam2023gpt}) for reasoning. To compensate for the lack of visual information, these methods further integrate LLMs with vision–language models (VLMs), such as CLIP \cite{radford2021learning}, to facilitate keyframe selection and supply visual context to the LLM. However, such multi-component pipelines introduce additional system complexity and still rely heavily on external retrieval modules, but also risk discarding fine-grained visual cues essential for answering detailed queries.
Another line of work employs LLM-based agents \cite{lin2023video,xu2024pllava,wang2024internvideo2,wang2024tarsier} (\cref{fig:intro}~(b)). instead pretrains a video based MLLM on large-scale multimodal datasets. Although effective, this paradigm requires extensive multimodal data and substantial computation, and often relies on many input frames to cover complex content, leading to high memory cost, redundancy, and low reasoning efficiency.

\captionsetup[subfigure]{font=small}

Recent progress in MLLMs for downstream tasks \cite{xiao2024interaction2code,shao2025growing,kim2024image} encourages rethinking how to efficiently exploit multimodal inputs for long-form video understanding. A simple baseline is to uniformly sample frames and feed them into an MLLM. This baseline often fails to capture critical events, while dense sampling drastically increases memory and computation. IG-VLM \cite{kim2024image} reduces memory by synthesizing multiple frames into a single image grid, but such compression does not explicitly target query-relevant frames for detailed reasoning. Overall, there is a need for an MLLM-based framework that performs query-driven video analysis using only a small number of frames.


Understanding long videos is analogous to reading comprehension: humans first locate the relevant paragraph and then focus on details within it. Inspired by this, we propose a \textbf{Pro}gressive \textbf{V}ideo \textbf{C}ondensation \textbf{A}gent, ProVCA, for long video question answering (\cref{fig:intro}~(c)). 
ProVCA is a progressive video condensation agent designed for efficient long-video understanding with MLLMs, following a top-down, progressively refined sampling process.
First, Segment Localization invokes the MLLM’s semantic reasoning capabilities over sparsely sampled global frames, enabling coarse-grained temporal navigation to identify the region most relevant to the query. Next, Snippet Selection partitions this region into semantically coherent snippets using caption-level similarity, and the MLLM evaluates snippet representations to filter out irrelevant temporal blocks. Finally, Keyframe Refinement returns to the visual domain within low-confidence snippets, where the MLLM jointly analyzes the full frame sequence to pinpoint the precise moments that are truly question-critical.
Through this hierarchical, progressively reasoning pipeline, ProVCA integrates semantic abstraction with fine-grained visual analysis, thus transforming long videos into a compact set of highly informative key frames. This design preserves strong MLLM reasoning accuracy while substantially reducing the computational cost, offering a scalable and cognitively motivated solution for long-video tasks.

We conduct extensive experiments on three long-form video understanding benchmarks: NExT-QA \cite{xiao2021next}, EgoSchema \cite{mangalam2023egoschema}, and IntentQA \cite{li2023intentqa}. ProVCA achieves zero-shot accuracies of 80.5\%, 69.3\%, and 77.7\%, respectively, surpassing the current training-free state-of-the-art method VideoTree \cite{wang2024videotree} by 4.9\%, 8.2\%, and 10.8\%. These improvements show that ProVCA can handle long videos with both higher precision and lower frame usage.

Our main contributions are summarized as follows:
\begin{itemize}
    \item We introduce a new paradigm for long video understanding that employs MLLM-based agents rather than pure-text LLM agents, enabling direct use of visual information and more accurate video reasoning.
    \item We propose ProVCA, a novel and effective framework that progressively operates at the segment, snippet, and frame levels to retrieve query-relevant content via top-down reasoning.
    \item Extensive experiments on multiple benchmarks demonstrate that ProVCA achieves a favorable balance between reasoning accuracy and efficiency, attaining state-of-the-art performance while using fewer frames.
\end{itemize}

\section{Related Work}

\subsection{Video Understanding Based on LLMs} 
The strong performance of LLMs in natural language processing has spurred their use in video understanding. Existing methods roughly fall into two categories. The first integrates LLMs with visual models into unified multimodal models via pre-training or fine-tuning \cite{lin2023video, li2024mvbench,shao2025imp}, which generally achieve strong performance but require large-scale multimodal instruction data and high computational cost.
A second category converts videos into text using pre-trained visual captioning models \cite{zhang2023simple, wang2024videotree}, enabling LLM-based reasoning but at the risk of discarding visual cues crucial for fine-grained questions.

With the emergence of MLLMs, recent works explore reasoning directly from video frames. IG-VLM \cite{kim2024image,yin2026videoarm,park2024too} concatenates multiple sampled frames into a grid image for MLLM inference, and LVNet \cite{park2024too} refines keyframes by applying visual templates before feeding frames into an MLLM. These approaches confirm that visual inputs can significantly strengthen video understanding, yet they still rely on uniform or heuristic sampling. In contrast, our method performs coarse-to-fine, query-aware frame selection, allowing the MLLM to focus on a compact but informative subset of frames.

\subsection{LLM-based Agents} 
LLM-based agents extend LLMs by invoking tools and modules and have been widely used in vision tasks. In video understanding, several agent frameworks dynamically call APIs and tools to enhance comprehension. VideoAgent \cite{wang2025videoagent} and DrVideo \cite{ma2024drvideo} iteratively retrieve key information from videos. TraveLER \cite{shang2024traveler} employs multiple LLM agents with different roles to jointly interpret complex videos.

Unlike these multi-component systems, we view long video understanding as a reading-comprehension-style task in the video modality. ProVCA progressively identifies and localizes relevant frames from coarse to fine granularity based mainly on MLLM reasoning, enabling a concise and effective pipeline without heavy reliance on external retrieval tools.

\section{Methodology}
In this section, we present the overall methodology, including preliminaries, the ProVCA framework, and each key component. We also explain how MLLM reasoning leverages these components to improve accuracy and efficiency.

\subsection{Preliminaries}\label{sec:pre}
We consider a video question answering task where a video $V = (v_1,...,v_n)$ and a question $Q$ are given, and the goal is to predict an answer $A$.

MLLMs extend LLMs by jointly processing text and visual inputs for cross-modal reasoning. A straightforward way to apply an MLLM is to uniformly sample frames from $V$ and feed them, together with $Q$, into the model:
\begin{equation}\label{eq:mback1}
A = f_{\text{MLLM}}(V, Q),
\end{equation}
where $f_{\text{MLLM}}$ denotes the MLLM. However, frames relevant to $Q$ are often sparsely and unevenly distributed, so uniform sampling tends to include many redundant frames and reduces reasoning efficiency.

To better exploit MLLMs, we reformulate the task as a two-stage process:
\begin{enumerate}
    \item \textbf{Frame Selection Stage:} The model identifies and selects a subset $V_s = \{v_j\}_{j=1}^{k}$ of $k$ frames from $V$ that are most correlated with $Q$, filtering out irrelevant frames.
    \item \textbf{Answer Prediction Stage:} The MLLM takes $V_s$ and $Q$ as input and predicts the final answer $A$.
\end{enumerate}
Formally, we can write:
\begin{align}\label{eq:mback2}
A = f_{\text{MLLM}}(V_s, Q), \quad s = f_{\text{MLLM}}(V, Q),
\end{align}
where $s$ denotes the indices of selected frames. ProVCA is designed to implement this two-stage formulation in a progressive manner.

\subsection{The ProVCA Framework}\label{sec:overview}
\begin{figure*}[t]
    \centering
    \includegraphics[width=1.0\textwidth, height=0.35\textwidth]{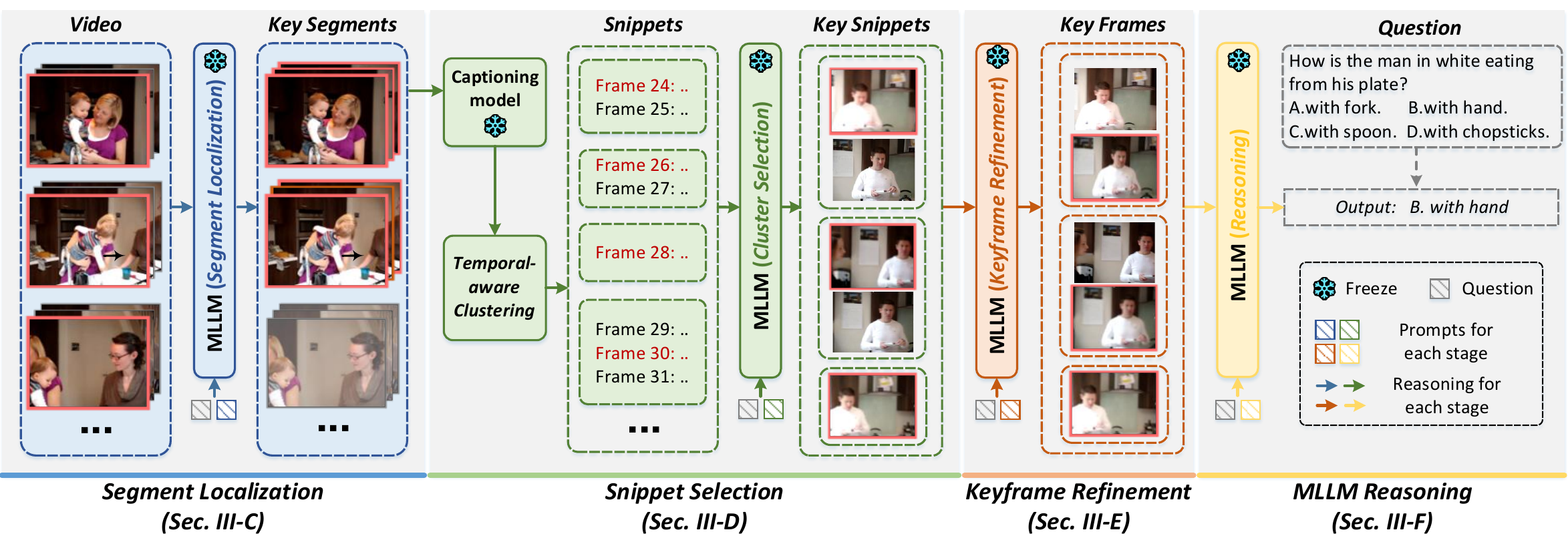}
    \vspace{-0.2in}
    \caption{Overview of ProVCA in video understanding based on MLLM. }
    \vspace{-0.25cm}
    \label{framework}
\end{figure*}

ProVCA is a Progressive Video Condensation Agent built on the reasoning ability of MLLMs. It operates in three stages to extract keyframes for a given query: (1) \emph{Segment Localization} (\cref{sec:segLoc}) scans the video to locate a relevant segment; (2) \emph{Snippet Selection} (\cref{sec:cluSelect}) divides that segment into snippets and selects those most relevant to the query; and (3) \emph{Keyframe Refinement} (\cref{sec:keyfrmRef}) further refines the selected snippets to identify keyframes. The final keyframes are then used for MLLM reasoning (\cref{sec:res}). \cref{framework} gives an overview of the entire process.

\subsection{Segment Localization}\label{sec:segLoc}  
To localize a coarse temporal region, we divide the video into $n$ segments and take the starting frame of each segment as input to the MLLM. These $n$ frames provide global context and enable the model to identify which segment is most related to the query. Concretely, we feed $(V, q)$, where $V$ denotes the sampled frames and $q$ the query, into the MLLM and obtain the segment
\(
S = (v_s, \dots, v_{s+m})
\)
that best aligns with $q$. This step narrows the search interval and reduces subsequent computation.

\subsection{Snippet Selection} \label{sec:cluSelect}  
Directly feeding all frames in $S$ into an MLLM remains costly. Instead, we first use captions derived from frames for efficient filtering. Existing agent-based methods \cite{wang2025videoagent, ma2024drvideo} often use additional modules such as CLIP for keyframe selection. In contrast, we rely primarily on MLLM reasoning, with captions used to reduce redundancy before visual reasoning.

A naive strategy would convert all frames in $S$ into captions and treat them independently. However, adjacent frames typically exhibit minimal differences and yield highly similar captions, which introduces redundancy and makes it harder to identify keyframes. We therefore group frames into snippets before querying the MLLM.

\vspace{3pt}
\noindent\textbf{Temporal-aware clustering.}  
We apply a captioning model to each frame in $S$ to obtain captions, and compute cosine similarity between their text embeddings. Starting from the first frame of $S$, we compare each subsequent frame with the first frame in the current snippet. If the similarity remains high, the frame is assigned to the same snippet; once it falls below a threshold, we start a new snippet. This procedure exploits both semantic similarity and temporal order.

To accommodate different segment lengths, we define a dynamic threshold $t$ based on the segment length $\left | S \right |$:
\begin{equation}
t = t_{max} - \frac{\left | S \right |-1}{\left | V \right |}  \times (t_{max} - t_{min}),
\end{equation}
where $t_{max}$ and $t_{min}$ are upper and lower bounds. Frames with similarity above $t$ remain in the current snippet, while those below $t$ initiate a new snippet. This yields $p$ snippets, each represented by its first frame.

\vspace{3pt}
\noindent\textbf{Key snippet selection.}  
We then input the representative frame of each snippet, together with the question, into the MLLM. The MLLM assesses the relevance between the snippet’s description and the query, and returns the indices of relevant representative frames along with confidence scores. A confidence of 1 indicates that the representative frame is likely sufficient for answering the question, whereas a confidence of 0 suggests that additional frames from the snippet should be examined in the refinement stage.

\subsection{Keyframe Refinement} \label{sec:keyfrmRef}
Based on snippet confidence, ProVCA adaptively decides whether refinement is needed. If a snippet has a confidence 1, its representative frame is directly used as a keyframe. Otherwise, we perform keyframe refinement within that snippet.

Captions alone may be insufficient to distinguish subtle differences between frames in a snippet, so we revert to the original images. All frames in the selected snippet are fed into the MLLM in chronological order, using a structured prompt indicating the snippet ID and the mapping between snippet and frame indices. This allows the MLLM to jointly analyze visual content in context and identify the frame most relevant to the query.

Compared with VideoTree \cite{wang2024videotree}, which repeatedly clusters many frames into a tree structure, our refinement focuses on selecting the single most relevant frame within each snippet, leading to a simpler and more targeted process.

\subsection{MLLM Reasoning}\label{sec:res}
Finally, the identified keyframes from all selected snippets are sorted chronologically and passed to the MLLM along with the query. We adopt a chain-of-thought prompting strategy \cite{wei2022chain} to encourage step-by-step reasoning. Because ProVCA has already filtered the video down to a small set of informative frames, the MLLM can devote its capacity to reasoning rather than scanning redundant content. This progressive localization significantly improves accuracy compared with uniform sampling while reducing computational cost.

\section{Experiments}
We evaluate ProVCA on three long-form video QA datasets: NExT-QA \cite{xiao2021next}, EgoSchema \cite{mangalam2023egoschema}, and IntentQA \cite{li2023intentqa}. We compare with state-of-the-art methods and conduct ablation studies to analyze each component.

\subsection{Datasets}
    \textbf{NExT-QA} contains 5{,}440 natural videos depicting everyday interactions, with an average length of 44 seconds. Questions are categorized as temporal, causal, or descriptive. We follow prior work and evaluate on a validation set of 570 videos and 5{,}000 multiple-choice questions, and use a subset of 500 QA pairs uniformly sampled across categories for ablations.
    
    \textbf{EgoSchema} is a large-scale benchmark for very long-form video understanding, consisting of over 5{,}000 egocentric videos (average 180 seconds) and over 5{,}000 multiple-choice QA pairs. We report results on a 500-Q subset and on the full test set via the official leaderboard.
    
    \textbf{IntentQA} focuses on intent reasoning, with 4{,}303 videos (average 44 seconds) and around 16{,}000 multiple-choice questions. We conduct zero-shot evaluation on its 2{,}000-question test set.

\subsection{Implementation Details} \label{sec:implementation}
    For EgoSchema and IntentQA, we decode videos at 1 fps, and for NExT-QA we uniformly sample 32 frames per video. Following VideoAgent, we use LaViLa \cite{zhao2023learning} as a captioner on EgoSchema and CogAgent \cite{hong2024cogagent} on NExT-QA and IntentQA. Text similarity between captions is computed using the ``sentence-transformers/all-MiniLM-L6-v2'' model with cosine similarity. Across all datasets, we employ GPT-4o as the MLLM, specifically the gpt-4o-2024-08-06 version, and use it in a zero-shot fashion without additional fine-tuning.

\subsection{Main Results}
    \cref{tab:main_results} compares ProVCA with recent state-of-the-art methods on NExT-QA, EgoSchema, and IntentQA. ProVCA is listed among methods that use proprietary MLLMs.
    
    \subsubsection{NExT-QA} 
    Compared with VideoAgent, VideoTree, and LVNet, ProVCA achieves an average accuracy of 80.5\%, improving over the strongest GPT-4-based baseline (VideoTree) by 4.9\%. Gains on the more challenging temporal and causal questions reach 4.9\% and 5.3\%, respectively. Relative to LVNet, which also uses GPT-4o, ProVCA improves performance by 7.6\%. It even slightly outperforms the large video-language model Tarsier-34B. These results are obtained while using only about 4.2 frames per video on average.
    
    \subsubsection{EgoSchema}
    On EgoSchema, ProVCA attains 69.3\% on the subset and 74.2\% on the full set, surpassing previous best methods by 8.2\% on the full set. ProVCA uses about 7.3 frames per video on average, compared with 12 frames for LVNet, showing that it is both more accurate and more efficient. Under the same GPT-4o and image-input configuration on the subset (\cref{tab:comparative_exper}), ProVCA still outperforms VideoAgent and VideoTree, confirming that the gain comes from the progressive condensation mechanism rather than model size or input modality alone.
    
    \subsubsection{IntentQA}
    On IntentQA, ProVCA reaches 77.7\% zero-shot accuracy, outperforming LVNet by 6.0\% while using the same MLLM. It processes only about 4.9 frames per video, substantially fewer than LVNet’s 12 frames. Since intent reasoning often requires aggregating cues from different parts of the video, these results suggest that ProVCA effectively gathers and condenses the necessary evidence for MLLM reasoning.
    
\begin{table*}[!ht]
\centering
\setlength{\tabcolsep}{5.5mm} 
\begin{tabularx}{\textwidth}{llcccccccccc}
\toprule
\multirow{2}{*}{\textbf{Method}} & \multirow{2}{*}{\textbf{LLM/MLLM}} & \multicolumn{4}{c}{\textbf{NExT-QA}} & \multicolumn{2}{c}{\textbf{EgoSchema}} & \multicolumn{1}{c}{\textbf{IntentQA}}  \\ 
\cmidrule(lr){3-6} \cmidrule(lr){7-8} \cmidrule(lr){9-9} 
 &            & \textbf{Tem.} & \textbf{Cau.} & \textbf{Des.} & \textbf{Avg.} & \textbf{Sub.} & \textbf{Full} & \textbf{Acc.}\\
\midrule
\multicolumn{9}{l}{\textit{methods based on open-source MLLMs}}\\

SeViLA\cite{yu2024self}         & Flan-T5 XL  & 61.3 & 61.5 & 75.6 & 63.6 & 25.7 & 22.7 & -\\
MVU\cite{ranasinghe2024understanding}         & Mistral-13B  & 55.4 & 48.1 & 64.1 & 55.2 & 60.3 & 37.6 & -\\
LangRepo \cite{kahatapitiya2024language}    & Mixtral-8×7B & 51.4 & 64.4 & 69.1 & 60.9 & 66.2 & 41.2 & 59.1\\
\midrule
\multicolumn{9}{l}{\textit{methods based on proprietary LLMs}}\\
LLoVi \cite{zhang2023simple}      & GPT-4        & 61.0 & 69.5 & 75.6 & 67.7 & 61.2 & - & 64.0\\
VideoAgent \cite{wang2025videoagent}  & GPT-4        & 64.5 & 72.7 & 81.1 & 71.3 & 60.2 & 54.1 & -\\
VideoTree \cite{wang2024videotree} & GPT-4        & 70.6 & 76.5 & 83.9 & 75.6 & 66.2 & 61.1 & 66.9\\
VideoAgent \cite{fan2025videoagent} & GPT-4        & - & - & - & - & 62.8 & 60.2 & - \\
DrVideo \cite{ma2024drvideo} & GPT-4         & - & - & - & - & 66.4 & 61.0 & -\\
TraveLER \cite{shang2024traveler} & GPT-4         & 70.0 & 60.5 & 78.2 & 68.2 & - & 53.3 & -\\
\midrule
\multicolumn{9}{l}{\textit{methods based on proprietary MLLMs}}\\
IG-VLM \cite{kim2024image}  & GPT-4V   & 63.6 & 69.8 & 74.7 & 68.6 & 59.8 & - & 65.3\\
LVNet \cite{park2024too}  & GPT-4o  & 65.5 & 75.0 & 81.5 & 72.9 & 68.2 & 61.1 & 71.7\\
\midrule
\textbf{ProVCA} & GPT-4o & \textbf{75.5} & \textbf{81.8} & \textbf{86.9} & \textbf{80.5} & \textbf{74.2} & \textbf{69.3} & \textbf{77.7}\\
\bottomrule
\end{tabularx}
\caption{Comparison with other training-free methods on EgoSchema, NExT-QA, and IntentQA.}
\label{tab:main_results}
\end{table*}

\begin{table*}
\renewcommand\arraystretch{1.2}
\begin{subtable}[t]{0.3\textwidth}
\centering
\begin{tabular}{lcc}
\toprule
\textbf{LLM} & \textbf{Model Size} & \textbf{Accuracy} \\
\midrule
LLaVA-1.6 \cite{liu2024llavanext} & 34B & 67.6 \\
LLaVA-OV \cite{li2024llava} & 72B & 79.8 \\
GPT-4V \cite{OpenAI2023} & - & 76.8 \\
GPT-4o \cite{hurst2024gpt} & - & \multicolumn{1}{>{\columncolor{mygray}}c}{\textbf{85.2}}\\
\bottomrule
\end{tabular}
\captionsetup{font=footnotesize}
\subcaption{\textbf{Capability of MLLMs.} More powerful models bring higher accuracy.}
\label{table:mllm}
\end{subtable}
\quad
\begin{subtable}[t]{0.3\textwidth}
\centering
\begin{tabular}{lc}
\toprule        
\textbf{Input}  & \textbf{Accuracy} \\
\midrule
captions & 73.0 \\
concat images & 81.2 \\
images & 84.2 \\
images + captions & \multicolumn{1}{>{\columncolor{mygray}}c}{\textbf{85.2}}\\
\bottomrule
\end{tabular}
\captionsetup{font=footnotesize}
\subcaption{\textbf{MLLM input.} Combining images with captions is most effective.}
\label{table:input}
\end{subtable}
\quad
\begin{subtable}[t]{0.36\textwidth}
\centering
\begin{tabular}{lcc}
\toprule
\textbf{Method} & \textbf{Accuracy} & \textbf{Frames} \\ 
\midrule
(a) w/o segment localization & 82.6 & 18.2 / 4.6 \\ 
(b) w/o snippet selection & 83.2 & 14.7 / 5.9 \\ 
(c) w/o keyframe refinement & 83.8 & 13.2 / 5.8\\  
(d) default & \multicolumn{1}{>{\columncolor{mygray}}c}{\textbf{85.2}} & 13.9 / 4.6 \\
\bottomrule
\end{tabular}
\captionsetup{font=footnotesize}
\subcaption{\textbf{Effect of ProVCA modules.} Each module improves both efficiency and accuracy.}
\label{table:module}
\end{subtable}
\\[1.em]
\begin{subtable}[t]{0.3\textwidth}
\centering
    \begin{tabular}{cc}
        \toprule
        \textbf{Segment Number} & \textbf{Accuracy} \\
        \midrule
        3 & 81.6 \\
        4 & \multicolumn{1}{>{\columncolor{mygray}}c}{\textbf{85.2}} \\
        5 & 83.6 \\
        6 & 82.4 \\
    \bottomrule
    \end{tabular}
\captionsetup{font=footnotesize}
\subcaption{\textbf{Effect of segment number.} Too few or too many segments both hurt performance.}
\label{table:frn}
\end{subtable}
\quad
\begin{subtable}[t]{0.3\textwidth}
\centering
\begin{tabular}{lc}
\toprule
\textbf{Method} & \textbf{Accuracy} \\
\midrule
(a) non-seq. (visual)  \cite{wang2024videotree} & 83.0 \\
(b) non-seq. (caption) & 82.2 \\
(c) seq. clustering (visual) & 82.8 \\
(d) seq. clustering (caption) &\multicolumn{1}{>{\columncolor{mygray}}c}{\textbf{85.2}} \\
\bottomrule
\end{tabular}
\captionsetup{font=footnotesize}
\subcaption{\textbf{Snippet clustering strategy.} Temporal caption similarity performs best.}
\label{table:div_strategy}
\end{subtable}
\quad
\begin{subtable}[t]{0.36\textwidth}
\centering
\begin{tabular}{lc}
\toprule
\textbf{Snippet Input } & \textbf{Accuracy} \\
\midrule
(a) independent (SeViLA \cite{yu2024self}) & 83.2  \\
(b) independent (MLLM) & 83.4  \\
(c) sequential &  83.2 \\
(d) global &\multicolumn{1}{>{\columncolor{mygray}}c}{\textbf{85.2}}  \\
\bottomrule
\end{tabular}
\captionsetup{font=footnotesize}
\subcaption{\textbf{Keyframe refinement.} Global refinement over snippets is most effective.}
\label{table:keyfrm_ref}
\end{subtable}
\caption{\textbf{Ablation experiments for ProVCA}. All results are on the val subset of NExT-QA. Best results are bolded, and default settings are highlighted in \colorbox{mygray}{gray}.}
\vspace{-0.5cm}
\label{table:abla}
\end{table*}

\begin{table}
\centering
\setlength{\tabcolsep}{3.8mm} 
\begin{tabular}{lcccc}
\toprule
\textbf{Method} & \textbf{MLLM} & \textbf{Input} & \textbf{Frames} & \textbf{Acc}\\ \midrule
VideoAgent \cite{wang2025videoagent} &  GPT-4o  &  Image & 7.7 & 73.2\\
VideoTree \cite{wang2024videotree} &  GPT-4o  &  Image & 6.4 & 72.4\\
ProVCA  & GPT-4o  &  Image & 7.1 & \textbf{74.2} \\ \bottomrule      
\end{tabular}
\caption{Comparison of ProVCA, VideoAgent, and VideoTree on the EgoSchema subset.}
\vspace{-0.5cm}
\label{tab:comparative_exper}
\end{table}

\subsection{Ablation Studies} \label{sec:ablation}
We conduct ablations on the NExT-QA validation subset to understand the impact of different design choices. The main observations from \cref{table:abla} are summarized below.

\subsubsection{MLLM choices} 
\cref{table:mllm} compares several open-source and proprietary MLLMs. GPT-4o achieves the highest accuracy (85.2\%), outperforming LLaVA-1.6, LLaVA-OV, and GPT-4V, which confirms that a stronger MLLM substantially benefits progressive reasoning. Among open-source models, LLaVA-OV performs best, likely due to its use of video-related data during instruction tuning.

\subsubsection{MLLM input}
\cref{table:input} examines different frame representations. Using only captions yields 73.0\% accuracy, while image-based inputs perform much better. Directly feeding multiple images surpasses concatenated image grids, and combining images with captions further improves performance to 85.2\%. This supports our design that uses images as the main MLLM input and captions only for snippet selection.

\subsubsection{Effect of ProVCA modules}
\cref{table:module} evaluates the contribution of each module. Removing segment localization increases the number of frames processed and reduces accuracy by 2.6\%. Disabling snippet selection and relying on uniform sampling reduces accuracy by 2.0\%. Without keyframe refinement, performance drops by 1.4\%. These results show that each stage—segment localization, snippet selection, and keyframe refinement—contributes to the overall accuracy–efficiency trade-off.

\subsubsection{Number of segments}
As shown in \cref{table:frn}, using four initial segments yields the best performance. Fewer segments provide insufficient temporal resolution, while more segments over-fragment the video and cause the MLLM to focus on overly narrow intervals, degrading performance.

\subsubsection{Snippet clustering strategy}
\cref{table:div_strategy} compares our temporal caption-similarity clustering with alternative strategies. Non-sequential k-means clustering, either on visual or caption features, underperforms our method, as it ignores temporal order and requires pre-defining the number of clusters. Sequential clustering based on visual similarity is better, but sequential caption-based clustering achieves the highest accuracy, consistent with our use of captions in snippet selection.

\subsubsection{Keyframe refinement} 
\cref{table:keyfrm_ref} investigates several refinement strategies. Independent selection within each snippet, either using SeViLA’s Localizer or an MLLM, leads to suboptimal performance, as it ignores cross-snippet context. A sequential strategy that conditions on the previous snippet brings limited gains. Our global refinement approach, which jointly considers all selected snippets, achieves the best accuracy (85.2\%), indicating that leveraging a more complete view of candidate frames helps identify more relevant keyframes.

\vspace{-0.1cm}
\section{Conclusion}
We have presented ProVCA, a Progressive Video Condensation Agent for efficient long video understanding with MLLMs. ProVCA progressively extracts critical visual content at segment, snippet, and frame levels, enabling the MLLM to reason over a compact set of keyframes rather than entire videos. Extensive ablation studies and comparisons on three challenging benchmarks demonstrate that ProVCA outperforms existing state-of-the-art methods, including large video-language models, while using fewer frames. We hope ProVCA can serve as a strong baseline and inspire further work on combining multimodal large language models with efficient content selection for long video reasoning.
\vspace{-0.1cm}
\section{Acknowledgements}
This work was supported by the Zhejiang Provincial Natural Science Foundation of China under Grant No. LQN26F020053, in part by the National Natural Science Foundation of China under Grants No. 62422204, in part by the Fundamental Research Funds for the Provincial Universities of Zhejiang under Grant No. GK259909299001-040, in part by the Zhejiang Provincial Natural Science Foundation of China under Grant No. LRG26F020001, in part by the Key Research and Development Program of Zhejiang Province No. 2025C01026, in part by the Scientific Research Innovation Capability Support Project for Young Faculty.

\bibliographystyle{IEEEbib}
\bibliography{icme2026references}

\end{document}